# The perpetual motion machine of AI-generated data and the distraction of "ChatGPT as scientist"

Jennifer Listgarten, University of California, Berkeley

Abstract:

Since ChatGPT works so well, are we on the cusp of solving science with AI? Isn't AlphaFold2 suggestive that the potential of LLMs in biology and the sciences more broadly is limitless? Can we use AI itself to bridge the lack of data in the sciences in order to then train an AI? Herein we present a discussion of these topics.

## Commentary

As a longtime researcher at the intersection of AI and biology, for the past year I have found myself being asked questions about large language models for science. One goes something like this: "Since ChatGPT works so well, are we on the cusp of solving science with AI?" Alternatively, "Isn't AlphaFold2 suggestive that the potential of LLMs in biology and the sciences more broadly is limitless?" And inevitably: "Can we use AI itself to bridge the lack of data in the sciences in order to then train an AI?"

I do believe that AI—equivalently, machine learning—will continue to advance scientific progress at a rate not achievable without it. I don't think major open scientific questions in general are about to go through phase transitions of progress with machine learning alone. The raw ingredients and outputs of "science" are not found in abundance on the internet, let alone existing at all. Yet the tremendous power of machine learning lies in data. Lots of it.

A major distinguishing factor for the sciences, as compared to more traditional AI fields such as Natural Language Processing and Computer Vision, is the relative lack of publicly available data suitable for domains in the sciences (see Appendix). There are simply vastly fewer data existing in the sciences, and these are often siloed by academics and companies alike. Acquiring appropriate scientific data for AI typically requires not only highly trained humans, but also high-end facilities with expensive equipment, making for an overall expensive and slow endeavor compared to humans simply going about their day by adding to the vast trove of images, text, audio and video on the world wide web. Not to mention humans sitting at a computer, trivially labeling such modalities in the case of labeled data.

DeepMind researchers performed a fantastic feat, substantially moving the needle on the protein structure prediction problem. Protein structure prediction is a tremendously important challenge with actual and still-to-be-realized impact. It was also, arguably, the only challenge in biology, or possibly in all the sciences, they could have tackled successfully. There are three primary reasons for this that together made for an extremely rare event: i) the problem of protein structure prediction is easily defined quantitatively, ii) there already existed sufficient data for this narrowly defined problem—slowly and expensively collected over decades, to meaningfully train a complex, supervised model with, and iii) assessment by way of held out proteins whose structure we already know, yield a precise, yet human-interpretable, quantitative metric that should more or less translate to many common use cases. Very few problems in the sciences are lucky enough to have all three of these. In fact, the most interesting and impactful questions may not yet be formulated at all, let alone in a manner suitable for machine learning, or with existing suitable data, or even a way to readily generate suitable data for machine learning.

There is the hope that we can bridge scientific data gaps by using AI to generate synthetic data. But we simply cannot get something for nothing—fresh information must be injected into the system one way or another for there to be a win. Just as we cannot build a perpetual motion machine, we cannot generate new information in a trivial cycle of information processing. The topic of AI-generated "synthetic" data can be somewhat nuanced and technical but can be made intuitively accessible.

One might say that AlphaFold2 used synthetic data from the model itself to improve accuracy. However, AlphaFold2 used real, but unlabeled sequence data, a strategy classically known as semi-supervised learning, which has both a long history and theoretical underpinnings. Generally, such a strategy may or may not be helpful. That the specific instantiation of semi-supervised learning in AlphaFold2 used real protein sequences fed through the model to obtain AI-predicted labels, does not make these data entirely AI-generated or synthetic—they remain anchored on real protein sequences. Moreover, semi-supervised learning is not a magic bullet for the problem of lack of supervised data—it can only help so much. Anyone touting the use of AI-based synthetic data should be able to explain in clear terms where the new information is coming from, relative to how they plan to use it. If one cannot reasonably do so, it's almost certain that the procedure will not be useful. As my good friend and colleague, Alyosha Efros—an internationally renowned computer vision researcher—recounts "A few times each year, a physician with less than 2000 MRI images contacts me about using our generative models to generate more training data. Now I know which doctors not to go to."

One path to generating useful synthetic data is to integrate human knowledge. For example, we can augment a labeled protein structure data set by rotating protein structure labels in the original training data set by a random amount, while keeping the underlying sequences as they were, and adding these synthesized pairs to the training data. In doing so, one is encoding the human belief—in this case corresponding to physical reality—that a protein 3D structure is, essentially, the same structure, even if rotated in space. It is telling that one can entirely replace such a "data augmentation" strategy by instead encoding this belief directly into the architecture of the neural network—a testament to the fact that a trained model has a data equivalence. Jahanian *et al.*, put it elegantly: "Generative models can be viewed as a compressed and organized copy of a dataset" (Jahanian et al., 2022). Neither data augmentation nor symmetry-encoding architecture is necessarily the better strategy. Critically, both are founded on the same information—information that is readily identifiable. In both cases, a human injects information, either by way of manipulated (rotated), augmented data to which they are declaring an invariance, or by changing the model architecture to encode that same invariance.

Might we use synthetic data from one model to help train another, AI model? Indeed, we can use an auxiliary model, AI-based or otherwise (*e.g.*, biophysics-based) to generate synthetic data for another AI model. For example, we might use physics-based simulations (*e.g.*, molecular dynamics) to generate data for one AI model that otherwise has access to only a small amount of true, experimental data. We can use one AI model to help another, so long as the two models hold different information, either by way of the data they were trained on, or their inductive biases. For any strategy of generating data from one AI model to be useful to train another (or the same model), there cannot be a trivial cycle in the pipeline. For example, we cannot generate data from a generative model only to directly feed these generated data back into that same model with the same learning objective—doing so is analogous to trying to build a perpetual motion machine. For a data-generating procedure to be useful, any cycle must have feedback in it to inject new information, such as filtering those generated data with an external procedure, human- or machine-based. One classical strategy of combining different AI models together, that of ensemble learning, can be quite powerful, and provably so the more different the models are in their predictions. More differences mean more information. Of course, if the different information is sufficiently incorrect, this procedure might degrade rather than help performance. This lesson should be taken more broadly, as there truly is no free data lunch.

Will generative, and other machine learning models help us to make progress in the sciences? Undoubtedly, yes. They already are. A generative model is, fundamentally, a probability density model, because it is capturing the probability distribution of the data, shaped by our hands in the way of inductive biases, explicitly known or not. We can use generative models to "score" unseen data samples to see if they "belong" in the set of training data; we can extract learned representations from them that may themselves yield scientific insights or prove useful for downstream tasks; and, we can generate "new" samples from them. But we should not forget that those "new" samples, extracted representations, and scoring, all span only the raw information of bits that was used to train them in the first place. The hope is that the model gives us more useful and ready access to those bits of information.

As for ChatGPT and its text-based relatives, these will undoubtedly continue to provide a new generation of incredibly useful literature synthesis tools. These literature-based tools will drive new engines of profound convenience, previously impossible and only dreamed of, such as those providing medical diagnosis and beyond—also those not yet dreamed of. But they will not themselves, anytime soon, be virtual scientists. AI can help us understand the data we've collected, and with enough of it, to generalize from them within reason. It can also help us to decide what to measure, initially, or iteratively. But in order to probe the limits of current scientific knowledge, we need data that we don't already have, to answer questions we may not yet even know to ask. And for this, we'll just need to get back to the bench and do more experiments.

## Appendix

Regarding the use of the term "science"—this term itself is in the eye of the beholder, but for the sake of this commentary we include biology, chemistry and physics to be in scope; and leave out the more applied fields such as medicine.

With respect to the amount of data available in different fields, it's difficult to make detailed comparisons, but in broad strokes, we note the following. The open-access paired image-text dataset, LAION (https://laion.ai/blog/laion-5b/), has nearly 6 billion paired examples (Schuhmann et al., 2022, https://laion.ai/blog/laion-5b/), and the Common Crawl data set had, as of June 2023, around 3 billion web pages comprising roughly 400 terabytes, with billions of new pages added each month (https://commoncrawl.org/). In contrast, in the sciences: the number of protein sequences in UniRef as of May 2023 is approximately 250 million sequences and in the decade 2012-2022 went up roughly 150 million (https://www.ebi.ac.uk/uniprot/TrEMBLstats). Supervised data sets in the sciences include Open Catalyst 2022 which contains 62K DFT relaxations for oxides (Tran et al., 2022); Open Direct Air Capture 2023 which contains 38 million density functional theory calculations on 8,800 Metal Organic Framework materials (Sriram et al., 2023), and AlphaFold2 (Jumper et al., 2021) trained on ~170K proteins and their structures with an additional 350K unlabeled sequences from UniClust30 (https://deepmind.google/discover/blog/alphafold-a-solution-to-a-50-year-old-grand-challenge-in-biology/). As of 2022, there are 1663 RNA structures (Deng et al., 2023). ChemSpider contains 128 million chemical structures as of Nov 2023 (https://www.chemspider.com/).

In terms of the expense of protein structure data used for AlphaFold2, "The replacement cost of the entire PDB [Protein Data Bank] archive is conservatively estimated at ~US$20 billion, assuming an average cost of ~US$100 000 for regenerating each experimental structure" (Burley et al., 2023).

## Acknowledgements

Thanks to Tyler Bonnen, James Bowden, Jennifer Doudna, Lisa Dunlap, Alyosha Efros, Nicolo Fusi, Aaron Hertzmann, Hanlun Jiang, Aditi Krishnapriyan, Jitendra Malik, Sara Mostafavi, Hunter Nisonoff and Ben Recht for helpful comments on this piece as it was taking shape.

## References

Burley, S. K., Bhikadiya, C., Bi, C., Bittrich, S., Chao, H., Chen, L., Craig, P. A., Crichlow, G. V., Dalenberg, K., Duarte, J. M., Dutta, S., Fayazi, M., Feng, Z., Flatt, J. W., Ganesan, S., Ghosh, S., Goodsell, D. S., Green, R. K., Guranovic, V., … Zardecki, C. (2023). RCSB Protein Data Bank (RCSB.org): delivery of experimentally-determined PDB structures alongside one million computed structure models of proteins from artificial intelligence/machine learning. *Nucleic Acids Research*, *51*(D1), D488–D508. https://doi.org/10.1093/NAR/GKAC1077

Deng, J., Fang, X., Huang, L., Li, S., Xu, L., Ye, K., Zhang, J., Zhang, K., & Zhang, Q. C. (2023). RNA structure determination: From 2D to 3D. *Fundamental Research*, *3*(5), 727–737. https://doi.org/10.1016/J.FMRE.2023.06.001

Jahanian, A., Puig, X., Tian, Y., & Isola, P. (2022). Generative Models as a Data Source for Multiview Representation Learning. *International Conference on Learning Representations*. https://arxiv.org/abs/2106.05258

Jumper, J., Evans, R., Pritzel, A., Green, T., Figurnov, M., Ronneberger, O., Tunyasuvunakool, K., Bates, R., Žídek, A., Potapenko, A., Bridgland, A., Meyer, C., Kohl, S. A. A., Ballard, A. J., Cowie, A., Romera-Paredes, B., Nikolov, S., Jain, R., Adler, J., … Hassabis, D. (2021). Highly accurate protein structure prediction with AlphaFold. *Nature 2021 596:7873*, *596*(7873), 583–589. https://doi.org/10.1038/s41586-021-03819-2

Schuhmann, C., Beaumont, R., Vencu, R., Gordon, C., Wightman, R., Cherti, M., Coombes, T., Katta, A., Mullis, C., Wortsman, M., Schramowski, P., Kundurthy, S., Crowson, K., Schmidt, L., Kaczmarczyk, R., & Jitsev, J. (2022). LAION-5B: An open large-scale dataset for training next generation image-text models. *Advances in Neural Information Processing Systems*, *35*. https://arxiv.org/abs/2210.08402v1

Sriram, A., Choi, S., Yu, X., Brabson, L. M., Das, A., Ulissi, Z., Uyttendaele, M., Medford, A. J., & Sholl, D. S. (2023). *The Open DAC 2023 Dataset and Challenges for Sorbent Discovery in Direct Air Capture*. https://arxiv.org/abs/2311.00341v2

Tran, R., Lan, J., Shuaibi, M., Wood, B. M., Goyal, S., Das, A., Heras-Domingo, J., Kolluru, A., Rizvi, A., Shoghi, N., Sriram, A., Therrien, F., Abed, J., Voznyy, O., Sargent, E. H., Ulissi, Z., & Zitnick, C. L. (2022). The Open Catalyst 2022 (OC22) Dataset and Challenges for Oxide Electrocatalysts. *ACS Catalysis*, *13*(5), 3066–3084. https://doi.org/10.1021/acscatal.2c05426